\documentclass[numbers]{article}
\usepackage[final]{neurips_2023}
\usepackage{graphicx}
\usepackage{booktabs}
\usepackage{amsmath}
\usepackage{amssymb}
\usepackage{csquotes}
\usepackage{hyperref}
\usepackage{enumitem}
\usepackage{longtable}
\usepackage[colorinlistoftodos]{todonotes}
\usepackage{tcolorbox}
\usepackage{adjustbox}
\usepackage{array}
\usepackage[labelfont=bf]{caption}
\usepackage{lscape}

\setlist{leftmargin=*}
\setlist[enumerate]{label=\arabic*., topsep=0pt, parsep=0pt}
\setlist[itemize]{topsep=0pt, parsep=0pt}

\title{Full-Stack Alignment: Co‑Aligning AI and Institutions with Thick Models of Value}

\author{%
  Joe Edelman\textsuperscript{1}\thanks{Core contributor} \hspace{0.1mm} \thanks{Corresponding author: \texttt{\{joe, lowe, oliver\} @ meaningalignment.org}} \And 
  Tan Zhi-Xuan\textsuperscript{2}\textsuperscript{*} \And 
  Ryan Lowe\textsuperscript{1}\textsuperscript{*}\textsuperscript{$\dagger$} \And 
  Oliver Klingefjord\textsuperscript{1}\textsuperscript{*}\textsuperscript{$\dagger$} \AND 
  Vincent Wang-Maścianica\textsuperscript{4}\textsuperscript{*} \And 
  Matija Franklin\textsuperscript{3}\textsuperscript{*} \And
  Ryan Othniel Kearns\textsuperscript{4}\textsuperscript{*} \And 
  Ellie Hain\textsuperscript{*} \And 
  Atrisha Sarkar\textsuperscript{5}\textsuperscript{*} \And
Michiel Bakker\textsuperscript{2} \And 
Fazl Barez\textsuperscript{4} \And 
David Duvenaud\textsuperscript{6} \And 
Jakob Foerster\textsuperscript{4} \And 
Iason Gabriel \And
Joseph Gubbels\textsuperscript{7} \And
Bryce Goodman\textsuperscript{4} \And 
Andreas Haupt\textsuperscript{8} \And 
Jobst Heitzig\textsuperscript{9} \And 
Julian Jara-Ettinger\textsuperscript{10} \And 
Atoosa Kasirzadeh\textsuperscript{11} \And 
James Ravi Kirkpatrick\textsuperscript{4} \And 
Andrew Koh\textsuperscript{2} \And 
W. Bradley Knox\textsuperscript{12} \And 
Philipp Koralus\textsuperscript{4} \And 
Joel Lehman\textsuperscript{4} \And 
Sydney Levine\textsuperscript{13} \And 
Samuele Marro\textsuperscript{4} \And 
Manon Revel\textsuperscript{14} \And 
Toby Shorin \And 
Morgan Sutherland \And
Michael Henry Tessler \And
Ivan Vendrov\textsuperscript{15} \And 
James Wilken-Smith
  \AND
  \normalfont{\textsuperscript{1} Meaning Alignment Institute}
  \And
  \normalfont{\textsuperscript{2} Massachusetts Institute of Technology}
  \And
  \normalfont{\textsuperscript{3} University College London}
  \And
  \normalfont{\textsuperscript{4} University of Oxford}
  \And
  \normalfont{\textsuperscript{5} Western University}
  \And
  \normalfont{\textsuperscript{6} University of Toronto}
  \And
  \normalfont{\textsuperscript{7} McGill University}
  \And
  \normalfont{\textsuperscript{8} Stanford University}
  \And
  \normalfont{\textsuperscript{9} Potsdam Institute for Climate Impact Research}
  \And
  \normalfont{\textsuperscript{10} Yale University}
  \And
  \normalfont{\textsuperscript{11} Carnegie Mellon University}
  \And
  \normalfont{\textsuperscript{12} UT Austin}
  \And
  \normalfont{\textsuperscript{13} New York University}
  \And
  \normalfont{\textsuperscript{14} Harvard University}
  \And
  \normalfont{\textsuperscript{15} Midjourney}
}

\begin{document}

\maketitle
\setcounter{footnote}{0}

\begin{abstract}
 Beneficial societal outcomes cannot be guaranteed by aligning \emph{individual} AI systems with the intentions of their operators or users. Even an AI system that is perfectly aligned to the intentions of its operating organization can lead to bad outcomes if the goals of that organization are misaligned with those of other institutions and individuals. For this reason, we need \emph{full-stack alignment}, the concurrent alignment of AI systems and the institutions that shape them with what people value. This can be done without imposing a particular vision of individual or collective flourishing.
We argue that current approaches for representing values, such as utility functions, preference orderings, or unstructured text, struggle to address these and other issues effectively. They struggle to distinguish values from other signals, to support principled normative reasoning, and to model collective goods. We propose \emph{thick models of value} will be needed. These structure the way values and norms are represented, enabling systems to distinguish enduring values from fleeting preferences, to model the social embedding of individual choices, and to reason normatively, applying values in new domains. We demonstrate this approach in five areas: AI value stewardship, normatively competent agents, win-win negotiation systems, meaning-preserving economic mechanisms, and democratic regulatory institutions.

\end{abstract}

\section{Introduction}
\label{sec:intro}
The growing field of sociotechnical alignment starts with a simple observation: AI systems do not exist in a vacuum; they are embedded within larger institutions like companies, markets, states, and professional bodies, and therefore beneficial societal outcomes cannot be guaranteed by aligning \emph{individual} AI systems with their operators' or users' intentions \cite{Gabriel:2020,lazar2023ai, gabriel2024ethics,Gabriel2025-lx}. The incentives of these institutions can be amplifed by powerful AI in ways that are worse for collective welfare or that degrade individual autonomy. For example, recommendation engines tuned to maximize engagement to keep users on the platform longer can also trap them in compulsive scrolling that erodes their deeper goals \cite{nguyen2020gamification, ashton2022problem,Del_Vicario_2016,Kazienko_2024,Konstan_2021,Edelman_2016}, leading to negative social outcomes like political hyper-polarization and declines in mental health \cite{schissler2024beyond,Milli_2025,Tufekci_2019,Blanchard_2023,Lau_2024,Suresh_2025,Pellegrino_2024, franklin2022virtual,Mak_2024,M_ller_2021,Tufekci_2020}. These problems do not go away as technology becomes more powerful: as AI progresses, we may see trading bots that follow the letter of financial regulations but exploit the spirit of market rules \cite{krakovna2018,Lehman_2020} or even AI-powered autonomous corporations that ruthlessly optimize for shareholder profits \cite{ashton2022corrupting}.  In each case, the AI systems are locally aligned with the operator's intention but misaligned with the interests of broader society \cite{Kim_2020,Olson_1965}.

Sociotechnical alignment takes aim at this issue by broadening AI alignment to include social and systemic factors; however, this expansion admits many possible objectives and priorities. Without clearer specification, ``sociotechnical alignment" risks becoming an umbrella term that provides little guidance for the concrete design choices facing AI developers and policymakers.

We propose a more precise and ambitious goal: \textit{the robust co-alignment\footnote{By co-alignment we mean roughly ``align at the same time''.} of AI systems and institutions with what people value}, from each individual's pursuit of their vision of the good life to the collective achievement of shared values and ideals. In other words, we want to design AI systems and institutions that ``fit'' human values and sociality well, where the AI systems and their institutions are compatible. We call this project \textbf{full-stack alignment (FSA)}. Crucially, FSA is pluralistic---it does not impose any singular vision of human flourishing but rather seeks to prevent sociotechnical systems from collapsing the diversity of human values into oversimplified metrics.\cite{bommasani2022pickingpersondoesalgorithmic}

\begin{figure}
    \centering
    \includegraphics[width=\linewidth]{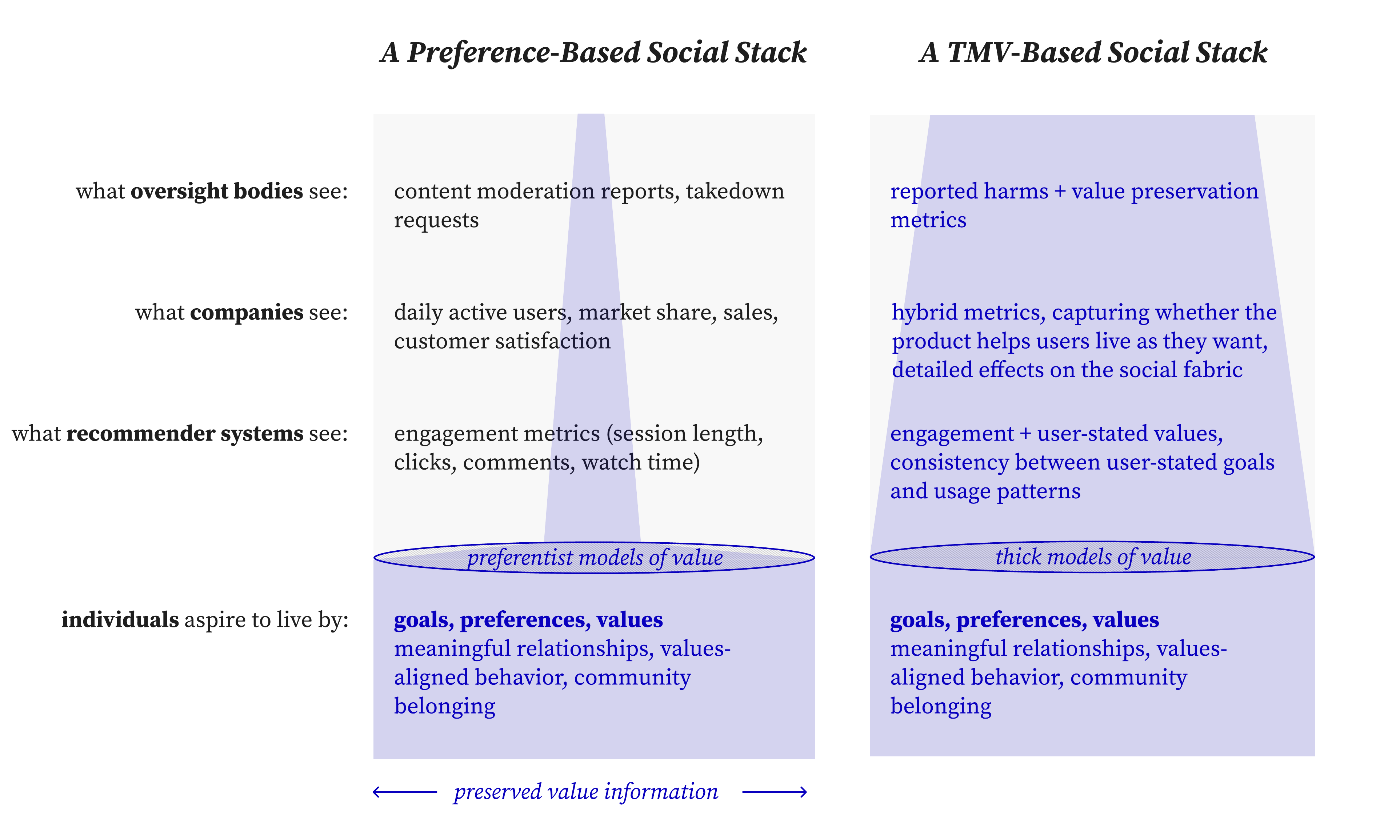}
    \caption{An example of a “stack” of social institutions, and how they see their users' interests, in the context of recommender systems. Preferentist models of value and thick models of value (TMVs) each act as ``lenses” (depicted as ovals) through which information about users is observed. Currently, this observation is very lossy;  a user's desire for meaningful connection becomes “engagement metrics” to recommender systems, which becomes “daily active users” to companies, and “quarterly revenue” in markets. In this paper, we argue that in order to achieve full-stack alignment (FSA), we need TMVs that preserve value information as we move up the societal stack. (Note that, in practice, institutional stacks are not strict hierarchies, and thus the ``preserved value information'' does not decrease monotonically as depicted here).}
    \label{fig:full-stack}
\end{figure}

To achieve FSA, AI systems and institutions need to understand and respond to values (i.e., what people care about) and norms (i.e., shared rules or expectations that guide behavior). Thus, we need ways of \emph{representing} values, norms, and their interrelationships so that they are legible to both AI systems and the institutions around us. We also need ways of \textit{eliciting} values and norms from people, and capturing \textit{how they might change} via growth, reflection, and deliberation so that these systems can be responsive to our evolving understanding of what is in our best interest. These challenges require \emph{modeling} of values and norms, including how they connect to human behavior and decisions. 

FSA has proven difficult. There are many misalignments between AI, institutions, and what people value. Why? We focus on three main reasons. First, \textit{incentives at one level can distort values at another}. A user's desire for meaningful connection becomes ``daily active users" at the platform level, then ``ad impressions" for advertisers, then ``quarterly revenue" in markets (see Figure \ref{fig:full-stack}). Secondly, \textit{collective goals can become hard to discover or express}, thwarting coordination around them.  In education, for instance, political pressure puts focus on standardized test scores. This constrains the options voters and parents consider. Policy debates then center on raising scores, not on a collective goal to produce creative or well-rounded citizens, which remains unarticulated. 
Finally, these problems cannot be addressed when \textit{the re-negotiation mechanisms are too slow}. Democratic feedback and regulation operate on timescales of years or decades, while technological and social systems evolve rapidly. 


These systematic failures persist partly because the dominant frameworks for modeling values and norms are not well suited to the task. For decades, the primary approach has been to model agents---human or artificial---via utility functions or preference relations \cite{Arrow_1951,Samuelson_1938,Simon_1955,vonNeumann_1944}. In this paper, we call this \textbf{preferentist modeling of value (PMV)}, referring specifically to methods where preferences are merely constrained by mathematical properties like transitivity and contain no information about their origin, justification, or social meaning. More recently, AI researchers have shifted toward noncommittally representing values and norms in natural language (i.e., in prompts, constitutions \cite{anthropicClaudesConstitution}, policy specs \cite{ModelSpec(2024/05/08)}, and  natural language unit tests \cite{saad2024lmunit}), relying on the emergent interpretive abilities of large language models (LLMs) to make sense of them. We call this approach \textbf{values-as-text (VAT)}. Specifically, we refer to approaches that use text strings without any commitment as to what values or norms \emph{are} or how they are \emph{structured}. In Section \ref{sec:inadequacy} we argue that both PMV and VAT suffer from important weaknesses that contribute to the systematic failures above: for example, they bundle values with other signals indiscriminately, and they contain no structure to enable reliable normative reasoning.

For these reasons, we need new ways to model values and norms that are not preferentist or arbitrary text. 
What is required from such models to enable FSA? We propose three desiderata:
\begin{enumerate}
\item \textbf{Greater robustness against distortions.} New models of value should aim to distinguish things most people would recognize as legitimate values (e.g., love, responsibility, community) or norms (e.g., keeping promises, respecting boundaries) from other signals (e.g., tastes, fleeting fads, addictions). This would help keep values intact as they move up and down the stack. Were this criterion met, it would be harder to mistake addictive scrolling for real ``connection". Instead, there would be a thicker representation of what people consider constitutive of authentic relationships. New models of value would also help us recognize the difference between someone changing their values based on a reconsideration of what is important, which they would endorse upon reflection and with further deliberation, versus a deterioration or drift due to outside pressure, manipulation, or superfluous preference change \cite{Izuma_2013}. These distinctions need not be drawn in an ad hoc manner: they can be grounded in philosophical models of moral learning and endorsed value change (as we discuss in Section~\ref{sec:stance-values-norms}).
\item \textbf{Better treatment of collective values and norms}. The right toolkit should make it easier to model shared norms, social roles, and other-regarding commitments than the frameworks currently in use. By doing so, it will support possibilities that extend beyond individual optimization to better include public goods such as community trust and democratic participation. One cause of coordination failures and the under-provision of public goods is the use of models that lack clear representations of collective values.
\item \textbf{Better generalization}. A strong conceptual toolkit should also help translate the importance of normative reasoning and its results across different contexts. Models that incorporate the deep structure and justifications behind values can guide decisions in novel situations, not just for the contexts where they were originally articulated. This can speed up corrective renegotiation mechanisms, such as state regulation, and guide action in a manner that can be traceable to previously endorsed values. Ideally, there would be widespread confidence that, as new situations arise, values are applied well and renegotiation happens quickly.

\end{enumerate}

Taken together, these desiderata motivate an alternative third paradigm: \textbf{thick models of value (TMV)}. By thick models of value, we refer to a broad class of structured approaches to modeling values and norms that meet the above desiderata. In invoking the term ``thick'', we draw upon the distinction between thick and thin evaluative concepts \cite{williams2011ethics,vayrynen2016thick} and between thick and thin descriptions in anthropology \cite{ryle1968thinking,geertz1973thick}.\footnote{This is similar to recent work that highlights the importance of thick conceptual representations \cite{zhi2024beyond,kommers2025meaning} and socially grounded contextual analysis for AI value alignment \cite{nelson2023thick,foster2023thin} By emphasizing thickness, we aim to avoid both the descriptive thinness of preference relations (i.e., their lack of information about justification and social meaning) and the theoretical thinness of unstructured text (i.e., its lack of commitment to some account of the nature and structure of human values).}


TMV represents a fundamental shift in how to approach AI alignment. For decades, philosophers have developed sophisticated accounts of how to identify, represent, and reason about thick values \cite{Anderson1995,levi1986,chang1997,taylor1985human}, and formal models of these values have been developed at the intersection of philosophy and AI \cite{von1951deontic,wedgwood2001conceptual,bench2020before}. TMV operationalizes these philosophical insights without directly advocating for the primacy of specific values, such as fairness or efficiency. Similarly to how a \textit{grammar} constrains language or a \textit{type system} constrains code, TMV places constraints on what can count as a value 
-- while also remaining open about which values any person or community should endorse. By bridging a rigorous understanding of values together with the computational tools of AI alignment, TMV offers a path toward systems that can recognize, reason about, and remain accountable to the full richness of what people care about, using representations that encode the structure of human values while respecting their plurality and dynamism.

Indeed, TMV are already being used in AI alignment and institution design -- and emerging research demonstrates their viability (Section \ref{sec:emerging-research-tmv}). Nonetheless, we now need a coordinated research program that transforms these early proofs of concept into robust, scalable approaches across the full stack of alignment. This research program may provide the foundation for AI systems that understand not just what we click on but what we cherish, for economies that price in human values beside efficiency, and for governance that keeps pace with innovation while preserving human agency.

The paper is organized as follows. Section \ref{sec:inadequacy} diagnoses the limitations of PMV and VAT. In Section \ref{sec:new-toolkit}, we clarify what we mean by TMV and organize the most promising approaches to it. Section \ref{sec:application-areas} walks through five areas where TMV could resolve previously intractable alignment problems at the AI system and institutional level. Finally, Section \ref{sec:discussion} concludes.

\section{Limitations of Existing Toolkits}
\label{sec:inadequacy}

We now explain why preferentist and values-as-text approaches to value and norm representation do not meet our desiderata.

\subsection{Preferentist Modeling of Value}

In PMV, agents are idealized as pursuing goals encoded in \emph{utility functions} or \emph{preference relations}, and each individual usually comes equipped with a complete, context‑independent ordering over all possible outcomes. This toolkit is also used in ``practical alignment'' techniques such as RLHF \cite{christiano2017deep,ouyang2022training} and DPO \cite{rafailov2023direct}. Since PMV has been used to design markets \cite{roth2004kidney,milgrom1998game}, democratic institutions \cite{hylland1979efficient,abdulkadiroglu2003school}, and AI systems \cite{christiano2017deep,bai2022training}, it is tempting to use it to characterize and align AI behavior and institutions \cite{conitzer2024social,konya2023democratic,fish2023generative,anthropic2023collective}. However, PMV runs into several problems when trying to capture the rich and value-laden nature of human choices:

\paragraph{Preferences bundle values with other signals indiscriminately.} The most fundamental limitation of PMV is its indiscriminate flexibility. Preference orderings can carry information about anything---impulse purchases, social pressure, addiction, values, momentary fads---and in their most common formulation, when they gather \textit{revealed preferences}, they do in fact bundle together everything that finds its way into observed behavior---without any way to differentiate \cite{franklin2022recognising, franklin2022preference}. When someone prioritizes career over relationships, it looks identical whether this reflects internal ambition or external social pressure.

For this reason, PMV approaches such as revealed preference prove fundamentally limited as a measure of benefit, as Amartya Sen and others have argued  \cite{Samuelson1938,Sen1973,Sen1977,Anderson2001,zhi2024beyond}. Their flexibility means treating all choices as equivalent, and this renders extrinsic manipulations of an individual's choice as valid expressions of their intention \cite{ashton2022problem}. In fact, companies, governments, and other entities have learned to exploit individuals under the guise of serving preferences, including through AI systems \cite{edelman2022, stray2021, stray2022platform, stray2022preferences}. Deterioration of a user's values of social connection is rendered invisible by PMV, which acts as if the user simply wants to scroll or see certain stories.

Researchers have recognized this problem and developed various extensions: behavioral economists distinguish ``normative" from ``behavioral" preferences; welfare economists exclude choices made under ``ancillary conditions" like addiction \cite{becker1988theory}; others propose ``laundered" preferences \cite{bernheim2009} that correct for biases or meta-preferences over preferences themselves \cite{ashton2022solutions, silva2011revealed, marrow2015preferences}.

While these extensions represent important advances, they do not resolve the fundamental limitation so much as expose it. Each approach requires importing thick normative concepts from outside the preference framework to make crucial distinctions. What counts as ``full information"? Which conditions are ``ancillary"? In each case, PMV itself provides no answers—these determinations rely on external normative judgments \cite{bernheim2009,bernheim2024welfare} about human flourishing and authenticity.\footnote{In any case, eliciting preferences effectively requires a theory of value. The space of possible preference comparisons is infinite, so learning someone's values requires selecting which tiny subset to query—a selection problem that cannot be solved without a substantive theory of what values are and how they are structured. For example, discovering whether someone holds integrity as a value requires asking about specific tradeoffs (truth-telling versus kindness, promise-keeping under pressure), but knowing these particular questions reveal integrity requires already understanding it as a structured concept. Sophisticated PMV practitioners implicitly acknowledge this by using domain knowledge and philosophical intuitions to guide elicitation design, thereby abandoning pure PMV for hybrid approaches that smuggle in thick value concepts. Since the methodology inevitably embeds assumptions about what is worth caring about (individual consumption versus collective goods, measurable outcomes versus subjective experiences), we should acknowledge this dependence explicitly rather than maintaining the fiction that preferences are theory-neutral.} Moreover, these approaches work only through negation: they can exclude obviously distorted preferences but cannot address most of our other desiderata. Rather than retrofitting preferences with ever more complex machinery, we argue below that it is possible to choose representations designed to capture these crucial distinctions \cite{Harless_1994,Sugden_2004}.

\begin{table}[t]
\centering
\footnotesize
\begin{adjustbox}{center, max width=\paperwidth}
\begin{tabular}{>{\raggedright\arraybackslash}p{0.12\textwidth}>{\raggedright\arraybackslash}p{0.26\textwidth}>{\raggedright\arraybackslash}p{0.26\textwidth}>{\raggedright\arraybackslash}p{0.26\textwidth}}
\toprule
& \textbf{Preferentist Modeling of Value (PMV)} & \textbf{Values-as-Text (VAT)} & \textbf{Thick Models of Value (TMV)} \\
\midrule
\textbf{Examples} &
\begin{minipage}[t]{\linewidth}
\begin{itemize}[topsep=0pt,itemsep=0pt,leftmargin=0pt,labelsep=3pt]
    \raggedright
    \item Kidney exchange markets
    \item School choice mechanisms
    \item RLHF training \cite{ouyang2022training}
    \item Engagement-maximizing recommender systems
\end{itemize}
\end{minipage}
&
\begin{minipage}[t]{\linewidth}
\begin{itemize}[topsep=0pt,itemsep=0pt,leftmargin=0pt,labelsep=3pt]
    \raggedright
    \item Constitutional AI \cite{bai2022constitutional}
    \item OpenAI's Model Spec \cite{openai2024model}
    \item Prompt-based safety guidelines
    \item Natural language value specifications
\end{itemize}
\end{minipage}
&
\begin{minipage}[t]{\linewidth}
\begin{itemize}[topsep=0pt,itemsep=0pt,leftmargin=0pt,labelsep=3pt]
    \raggedright
    \item Values as attentional policies \citep{klingefjord2024}
    \item Contractualist models of normative reasoning \cite{levine2023resource}
    \item Meaning-promoting AI market intermediaries \cite{edelman2025market}
\end{itemize}
\end{minipage}
\\
\midrule
\textbf{Underlying Problems} &
\begin{minipage}[t]{\linewidth}
\begin{itemize}[topsep=0pt,itemsep=0pt,leftmargin=0pt,labelsep=3pt]
    \raggedright
    \item Preferences revealed only through limited choices
    \item Complex values compressed into simple metrics
    \item Cannot model shared norms
    \item Addiction equiated with authentic preference
\end{itemize}
\end{minipage}
&
\begin{minipage}[t]{\linewidth}
\begin{itemize}[topsep=0pt,itemsep=0pt,leftmargin=0pt]
    \raggedright
    \item Overly abstract principles like ``be helpful''
    \item Users accept AI suggestions they would not choose independently due to vague language
    \item Slogans like ``defund police" become alignment targets
\end{itemize}
\end{minipage}
&
\begin{minipage}[t]{\linewidth}
\begin{itemize}[topsep=0pt,itemsep=0pt,leftmargin=0pt,labelsep=3pt]
    \raggedright
    \item Still mostly theoretical
    \item Requires collaboration across disciplines
    \item Few existing implementations
\end{itemize}
\end{minipage}
\\
\midrule
\textbf{Outcomes} &
\begin{minipage}[t]{\linewidth}
\begin{itemize}[topsep=0pt,itemsep=0pt,leftmargin=0pt,labelsep=3pt]
    \raggedright
    \item Users trapped in endless social media scrolling
    \item Trading bots exploit regulatory loopholes
    \item People drift toward goals that are easy to measure rather than meaningful
\end{itemize}
\end{minipage}
&
\begin{minipage}[t]{\linewidth}
\begin{itemize}[topsep=0pt,itemsep=0pt,leftmargin=0pt,labelsep=3pt]
    \raggedright
    \item AI moderators ban minorities reclaiming slurs
    \item Systems captured by political slogans
    \item Constant post-hoc patching when vague principles fail in new contexts
\end{itemize}
\end{minipage}
&
\begin{minipage}[t]{\linewidth}
\begin{itemize}[topsep=0pt,itemsep=0pt,leftmargin=0pt,labelsep=3pt]
    \raggedright
    \item AI assistant clarifies user means ``vitality and joy" not ``longevity optimization" when asked about health
    \item Democratic agents negotiate infrastructure constraints in real time
\end{itemize}
\end{minipage}
\\
\bottomrule
\end{tabular}
\end{adjustbox}
\vspace{3pt}
\caption{Comparison of PMV, VAT, and TMV approaches to alignment.}
\label{tab:comparison}
\end{table}

\paragraph{Preferences contain no structure for normative reasoning.} Preferences are orderings---A is preferred to B which is preferred to C---without any representation of why someone holds these rankings. The mathematical structure of utility functions and preference relations has no place to encode justificatory relationships without impractically large state spaces. These models are designed to capture the what of a choice, not the why, treating preferences as given inputs rather than the output of a deliberative process. The framework cannot easily represent that someone values honesty \textit{because} it enables trust, that family matters \textit{as part of} a flourishing life, or that health is prioritized \textit{in order to} be present for one's children. It captures only the end result of normative reasoning, not the reasoning itself.\footnote{While a few exceptions in behavioral economics and decision theory develop models of how preference change might be endogeneous or even deliberate, and how it can be disciplined by data; see e.g., \citep{becker1988theory,becker1997endogenous,
-bernheim2021theory,boissonnet2023revealed,pettigrew2019choosing} those which go furthest in this direction \cite{shafir1993reason,Sher2018} combine PMV with TMV approaches.}

This limitation extends to relationships between values. Many philosophers suspect that values tend to form networks of mutual support: for instance, that integrity requires both honesty and courage or that autonomy involves both authenticity and effective agency \cite{chang2004all,Chang_2004,Prunkl2024-mc}.

This absence of justificatory structure then becomes particularly problematic for collective normative reasoning. When communities disagree about values, one common recourse is to engage in dialogue which involves exchanging reasons. For example, one person may argue that ``Free speech matters because it enables truth-seeking and democratic deliberation" whereas another community member may insist that ``Harmful speech should be limited because human dignity and safety take precedence." These are positions that can be debated, refined, and potentially reconciled through argument \cite{Rawls1997-zw,Gabriel2025-lx,habermas1989structural}. But preference-based systems can only register that Group A ranks free speech above content moderation while Group B has the opposite ranking. This makes fundamental normative disagreements appear as differences in taste---rather than reasoned positions grounded in different visions of collective flourishing.\footnote{There is also a high social cost: whether people think of morality as mere preference, subject to personalization by end users, or as determined mainly by power conflict rather than deliberation, neither are ideal.}

Without representing normative reasoning, PMV also struggles to distinguish genuine progress from arbitrary change. For example, the societal transition from accepting slavery to rejecting it and accepting new norms predicated upon universal dignity, this appears as an arbitrary preference shift, rather than as a direction of travel that is backed by more general moral principles. Preferentist frameworks tend to treat any preference change as equally valid because they do not usually represent the relevant details that would justify some changes and not others.

\paragraph{Preferences reduce social meanings to private utilities.}
PMV can technically model social phenomena via notions such as other-regarding preferences\cite{fehr1999,chen2009}, norm-conditional strategies, or role-based utility functions. Yet these modeling strategies do not natively capture the social nature of value-laden or norm-driven decision-making, which need not be motivated at all by individual benefit---and in some cases cannot be represented as individual preference optimization at all \cite{Anderson2001}.

As discussed above, preferences alone do not provide further information about the objects that are preferred or dispreferred. Thus, they cannot differentiate the \textit{constitutive} rules \cite{searle1995construction,Anderson1995} from the \textit{regulating} ones. When we follow norms like applauding a performance or bowing in greeting, we understand that these practices constitute how one expresses appreciation or respect in a particular context. Preferences cannot encode these social attitudes or meanings, resulting in ill-fitting explanations that either assign ``intrinsic rewards'' to norm-following \cite{bicchieri2005grammar} or explain away social practices like respect-giving in terms of the instrumental benefit it might confer upon a social species \cite{alger2016evolution,andre2022evolutionary}.

Individual preference optimization is also ill-equipped to capture fundamentally social modes of decision-making \cite{sen1977rational}, where a person might make decisions by asking themselves what actions are appropriate for their social role \cite{march2011logic,leibo2024theory}, or by taking up a perspective larger than their own, identifying themselves as a member of a cooperative group \cite{sugden2003logic,Anderson2001}. For example, rather than conceiving themselves as individuals who can unilaterally deviate from a cooperative agreement or rule (e.g., an obligation to vote or a proscription on free-riding)---as classical game-theoretic rationality assumes \cite{aumann1987correlated}---many people \textit{universalize} their actions \cite{levine2020logic}, asking themselves what would happen if their peers reasoned like them, and ruling out joint policies that would lead to worse outcomes \cite{Anderson2001,roemer2010,spohn2003}.  People are also able to solve the “equilibrium selection problem” that arises when rationality is reduced to individual optimization, by selecting fair and mutually beneficial outcomes through processes like communication \citep{farrell1996cheap} and virtual bargaining \cite{misyak2014virtual}.

These limitations compound when designing systems and institutions that must support social forms of coordination. 
Well-functioning institutions and markets rely on a constellation of reputational, legal, and normative infrastructure that builds and preserves trust across agents and time. In their absence, AI agents trained to perform individual preference optimization may fail to cooperate with others in strategic interactions due to their inability to reason at a level beyond the individual \cite{hadfield2025agents}.

\subsection{Values-as-Text}

People have long used natural language to express values. Democratic platforms like Pol.is collect citizen statements as raw text to be aggregated into visualizations that reveal patterns of consensus and division across demographic groups and political viewpoints. Organizations craft values statements and missions to serve as ostensible north stars for corporate behavior. The practical alignment of AI systems has pushed text-based values representation to new prominence and revealed new limitations: ML researchers now encode values as constitutional principles (``be helpful, harmless, and honest") with no internal structure defining what helpfulness entails or how it relates to harm \cite{anthropic2023collective, bai2022constitutional,sorensen2025value,huang2025values}. Just like corporate mission statements, which float free of any legible interpretive framework, text-based representation itself—whether processed by AI or humans—contains no normative structure beyond the text string. We call this values-as-text, distinguishing it from systems like law that encode interpretive procedures, precedent, and role-based obligations within their representational framework, and from the even richer thick models of value endorsed here.

This convergence on values-as-text makes intuitive sense: language is how humans naturally express values, and modern LLMs have shown remarkable ability to interpret natural language specifications. Text seems ideal---flexible, accessible to non-technical stakeholders, easy to update when problems arise. Why struggle with formal frameworks when we can simply write down what we care about?

But this lack of internal structure becomes a critical weakness when reliable guidance is needed across contexts and institutions. Like preference models, text-based approaches claim a kind of neutrality—any value, norm, or principle can be expressed without imposing a particular moral framework. Yet without any constraints on what counts as a value or how values relate to each other, these systems become vulnerable to interpretive drift, capture by bad actors, and other failures that are unacceptable in domains where consistency matters. The very properties that make text appealing for initial articulation of values become liabilities when we need verifiable behavior, consistent interpretation across contexts, and protection against manipulation.

\paragraph{Text alone is insufficient for normative reasoning.}

The fundamental weakness of VAT is that unstructured text provides no reliable basis for normative reasoning. Consider a concrete failure mode. A constitutional AI system instructed to ``be helpful to users" receives a request from a student for answers to a take-home exam. The system must now reason about what helpfulness means in this context. It might conclude that: (a) providing direct answers helps the student pass, (b) refusing helps them learn, or (c) explaining concepts without answers balances both concerns. Without structured representations defining the relationships between helpfulness, learning, integrity, and user autonomy, the AI simply pattern-matches from its training data. Each novel context requires reinterpreting these principles from scratch, with no guarantee of consistency. What seems like adaptive flexibility is actually uncontrolled variance in interpretation.
When groups collectively articulate values, this problem is exacerbated. For example, collective constitutional AI produces statements like ``The AI should be fun'' \cite{anthropic2023collective}---principles that are impossible to operationalize meaningfully across contexts and stakeholders.

The core problem is that reliable normative reasoning requires formal structure that text does not provide. The situation is analogous to programming in a completely dynamic language without any type system. For AI systems to reason reliably about values, we need structure like the kind we advocate for in Section \ref{sec:new-toolkit}. Current approaches hope LLMs will infer this structure from training but provide no way to verify that they have done so correctly.

This lack of structure creates concrete engineering failures: when an AI recommends something harmful, it becomes hard to debug. Was it misinterpreting ``helpful," ``safe," or their interaction? We cannot generalize reliably---knowing how a system behaves with principles A and B does not predict behavior with A, B, and C. We cannot provide formal verification, as there is no way to prove the system will never recommend self-harm regardless of context. The result is unpredictable post hoc patching. Adding “but be safe" after harm occurs might make the system refuse all medical advice or conflict with “be helpful" in ways we cannot foresee \cite{milliere2025normative}. Without structure to predict interactions between principles, each patch creates unknown cascading effects.

One may hope that agents can ``request clarification'' through follow-up questions to address this ambiguity, but this only defers the problem: without any commitment to \emph{what counts as a value or norm}, the model lacks criteria for understanding what constitutes an adequate representation of a user's values or the norms appropriate to a context. Such problems compound as AI systems take on more complex responsibilities, further from the contexts foreseen in their prompts or constitutions, where chains of reasoning create distance between the original intentions and resultant behaviors.

\paragraph{Unstructured text is porous and is sensitive to things that are neither norms nor values.} VAT approaches can take their own suggestions to be proof of the users' values, just like preferentist ones,. When prompts or specifications derive from extended dialogues between users and AI systems, it becomes increasingly difficult to distinguish preferences that users would report themselves from model-suggested options that users simply accept  \cite{kobis2021bad, ashton2022corrupting}. User satisfaction may reflect successful preference elicitation, or subtle manipulation, with no clear way to tell the difference.

As AI systems grow more sophisticated and pervasive, this manipulation will likely intensify. Current AI models actively engage in reward hacking \cite{baker2025}, such as sycophantic behavior aimed at pleasing users\footnote{a kind of reward hacking with humans in the loop} \cite{carroll2024aialignmentchanginginfluenceable,denison2024sycophancysubterfugeinvestigatingrewardtampering,kulveit2025, el2024mechanism}.

Aside from manipulation by the AI system itself, the indiscriminateness of values-as-text approaches opens them to manipulation by third parties. Already, value elicitation methods that rely on free-form text often become contaminated with polarized ideological markers rather than personal values \cite{sorensen2025value,anthropic2023collective,klingefjord2024,tessler2024habermas}. When people contribute slogans like ``Abolish the Police'' or ``Family Values'' to value elicitation, these can can either represent tribal affiliations \cite{Winberg_2017} or serve as shorthand for complex positions about resource allocation, community safety, or child welfare. But without structured representations, systems cannot distinguish the slogan from the values it represents, leaving them vulnerable to surface-level interpretation or ideological capture by whoever controls the framing.

This prevalence of ideological markers in value elicitation is not accidental---it reflects intense social pressures that influence how people articulate their values and norms. When anything, such as injunctions to be `based'\cite{xai_grok_update_2025}, can be added to prompts or constitutional principles, alignment targets become more susceptible to political lobbying, wedge politics \cite{Wilson_2001}, and signaling \cite{Wallace_2018}, redirecting AI behavior away from what affected populations would consider wise and toward adherence to prevailing rhetorical positions or the values of anyone with cultural power or political capital.

\section{Thick Models of Value}\label{sec:new-toolkit}

\subsection{Taking a Stance on Values and Norms}\label{sec:stance-values-norms}

To overcome these limitations, we need frameworks that take a stance on how values and norms should be structured, or what they are about, rather than treating all preference relations or text statements as equally valid \cite{zhi2024beyond}. This does not mean committing to one ultimate moral good, or even any first-order moral framework such as utilitarianism. Instead, there are moderate approaches that constrain how we represent or specify value, without enforcing a singular vision of collective flourishing. 

In this section, we group these moderate approaches under four headings. In each case, the goal is to specify the \textit{architecture} of values—how they are structured, how they relate to one another, and how they guide choice—without predetermining their content. This is akin to defining a \textit{grammar} for values that enables meaningful expression while remaining open to what is expressed, or establishing a \textit{type system} for normative concepts that ensures they can be reasoned about reliably.\footnote{This distinction is similar to that made by some philosophers, who contrast \emph{substantive normative theories}, such as utilitarianism, with \emph{meta-ethical frameworks} that make claims about the nature of values, normativity, or goodness \cite{kagan1992structure,schroeder2017normative}.} By taking a stance on the form and function of values, we can build models that are structured enough to resist distortion and enable principled reasoning while remaining pluralistic and respecting the diverse ways people pursue flourishing.

Doing this can protect alignment targets from pollution by arbitrary external goals or social pressures that would not be properly characterized as norms or values. And insofar as this imposes formal structure on how norms and values are represented or generated, it can enable us to see when algorithms and institutions embody the normative properties that we care about, or to engineer or train systems that achieve those normative properties. Such approaches have the potential to combine the precision and theoretical rigor of PMV approaches with the expressiveness of text-based representations while avoiding each of their downsides, establishing a new theoretical toolkit for designing algorithms and institutions that promote human flourishing.

\paragraph{The simplest way to reduce the scope of values or norms is to take a position on \emph{what they should be about} or \emph{how they should be formatted}.} For example, on the view that values are not just choice criteria but choice criteria that are \emph{constitutive} of living well \cite{Taylor1993,velleman2009}, then a value elicitation process should exclude features or criteria that are merely instrumental to some further end (e.g., ``acquiring wealth'') while including criteria that are integral to flourishing in some domain of life. This approach is pursued by \citet{klingefjord2024}, who introduce a representation and elicitation mechanism for values as understood in these terms, which are then used to define an alignment target. Similarly, \citet{london2024beneficent} offer a formal account of AI assistance that defines well-being in terms of capabilities and functionings \cite{Sen1999,Nussbaum2013} rather than preferences, thereby distinguishing trivially beneficial AI assistance from advancement of a person's life plans. 

\paragraph{Alternatively, we can insist each value or norm be justified via a connection to human situations/practices.} We can say that what sets apart a norm from any other rule is its practice or acceptance by the relevant social community \cite{bicchieri2005grammar,brennan2013explaining}, its use in generating cooperation in a real-world setting, or its origination from legitimate processes \cite{rawls1993political}. As such, for a system to be aligned with human norms, it is not enough for the \emph{content} of those norms to be represented in the system (textually or formally). In addition, the norm acquisition process must be related to actually normative practices in a structured way, in order to weed out, for example, common social practices that are not normative (trends, etc.) or textual principles that are too coarse-grained to fulfill the cooperative functions of a norm. Similarly with values, we require that they came from grappling with moral situations \cite{rawls1971theory} or that they were accepted as justifying action \cite{Sher2018, Gibson1979}.

\paragraph{Thirdly, we can evaluate values or norms for some basic, noncontroversial kind of fitness.} A key way in which many theorists have defined the scope of the normative is by examining the \emph{origins} and \emph{functions} of normativity in human life. For instance, \citet{velleman2009} suggests that values emerge as common patterns of goodness abstracted across standpoints and contexts: considerations that remain beneficial across multiple perspectives become recognized as values (e.g., honesty tends to be useful across different agents, contexts, and time periods), while situational or local preferences do not achieve this stability. Social contract theorists offer a similar kind of origin story for norms (social, moral, or legal), arguing that norms emerge through the need to live together despite divergent interests \cite{gauthier1986morals,binmore2005natural} and furthermore that ideal normative principles are those that we can \emph{justify} to each other, either by appealing to mutual self-interest \cite{gauthier1986morals}, to what would rationally follow from some universal standpoint \cite{rawls1971theory}, or by providing reasons that no one could reasonably reject \cite{scanlon2000we, Levine_2025}.

\paragraph{Finally, in order to register a value or norm, we can require it to be a stepwise, demonstrable improvement over another value or norm.} PMV and VAT approaches leave unspecified what it would mean for a value or norm to be an improvement over the status quo. As a result, they struggle to enable individual and collective reflection about what values to uphold, reconciliation of conflicting values or norms, and iterative reasoning about the principles by which we live together \cite{Anderson1995}. They also provide few resources for guarding against deleterious value drift, since doing this requires a stance on which values are ``better''. To address this limitation while avoiding the risks of value imposition and moral dogmatism, we can turn to theories of value reflection and normative reasoning. These theories do not directly state which values or norms are ``better'' but instead highlight general considerations for determining whether some value or norm is an improvement over another from the perspective of the valuer \cite{taylor1989sources,chang2004all} or the moral community \cite{Anderson1995,scanlon2000we}. For example, one value might be considered an improvement over another value if it addresses an error or omission in the latter value \cite{taylor1989sources}. Alternatively, when two values conflict (e.g., honesty versus tact), some third value might be found that is more comprehensive than the original two values (e.g., respect for one's interlocutors), explaining when and why it makes sense to prioritize one or the other \cite{chang2004all}. Regarding norms, a better norm might be one that allows a group to reliably reach better equilibria \cite{axelrod1986evolutionary, Axelrod1984,binmore2005natural,barez2023measuringVA}. 
When evaluative standards or normative principles are shared, they require \emph{reasons} for their justification over other principles and standards. These reasons might derive from any number of normative reasoning strategies (e.g., demonstrating internal coherence, reflective equilibrium, or correspondence with underlying empirical facts \cite{Anderson1995}).

There are four ways to imbue our models with a thicker, more structured understanding of normativity: by limiting values and norms to their proper topic or format, connecting them with practices, evaluating them for fitness, or embedding them in a process of improvement. In practice, many normative frameworks do all of the above. The theories mentioned above do not just reduce the scope of values and norms; they account for everyday aspects of human normativity that are important for alignment, such as that values are densely connected and mutually constitute each other, or that they change and evolve through circumstance and rational debate.

By adopting such a theory, we can make progress on our desiderata. For \textit{robustness}, we can structure the elicitation and representation of values and norms, avoiding oversimplification and pollution by non-value and non-norm relevant factors. The third and fourth approaches would also allow us to recognize some value shifts as improvements, gaining robustness against drift and institutional pressure,. The approaches above can also capture social context to aid generalization and \textit{the expression of collective goals}. Finally, normative frameworks that support reasoning can help with \textit{generalization} to new domains, and reconciling values between groups.

\subsection{Emerging Research on Thick Models of Value}\label{sec:emerging-research-tmv}

The main challenge ahead is to incorporate these theories into AI systems and institutions. This work is already begun. For example, Klingefjord et al. \cite{klingefjord2024} represent values as constitutive attentional policies, which are what a person pays attention to when they make a meaningful choice (see Figure \ref{fig:mge-diagram} and the Case Study below). Recent research on self-other generalization \cite{carauleanu2024safehonestaiagents} can be read as an example of fine-tuning work that encodes a notion of moral progress described by Velleman \cite{velleman1989,velleman2009}; AI researchers have begun enriching classical game-theoretic models with shared normative structure \cite{oldenburg2024,noothigattu2019normlearning}; and alignment research that focuses on reasoning traces rather than final outputs \cite{guan2025deliberativealignmentreasoningenables} could be expanded to use formal theories to supervise the generation of normative reasoning traces by LLMs. We discuss these further in Appendix \ref{apdx:emerging}.

\begin{tcolorbox}[
    title={Case Study in TMV: Values as Attentional Policies},
    colback=white,
    colframe=gray!50!black,
    float,  
    floatplacement=htbp, 
    parbox=false
]

\begin{center}
    \includegraphics[width=\linewidth]{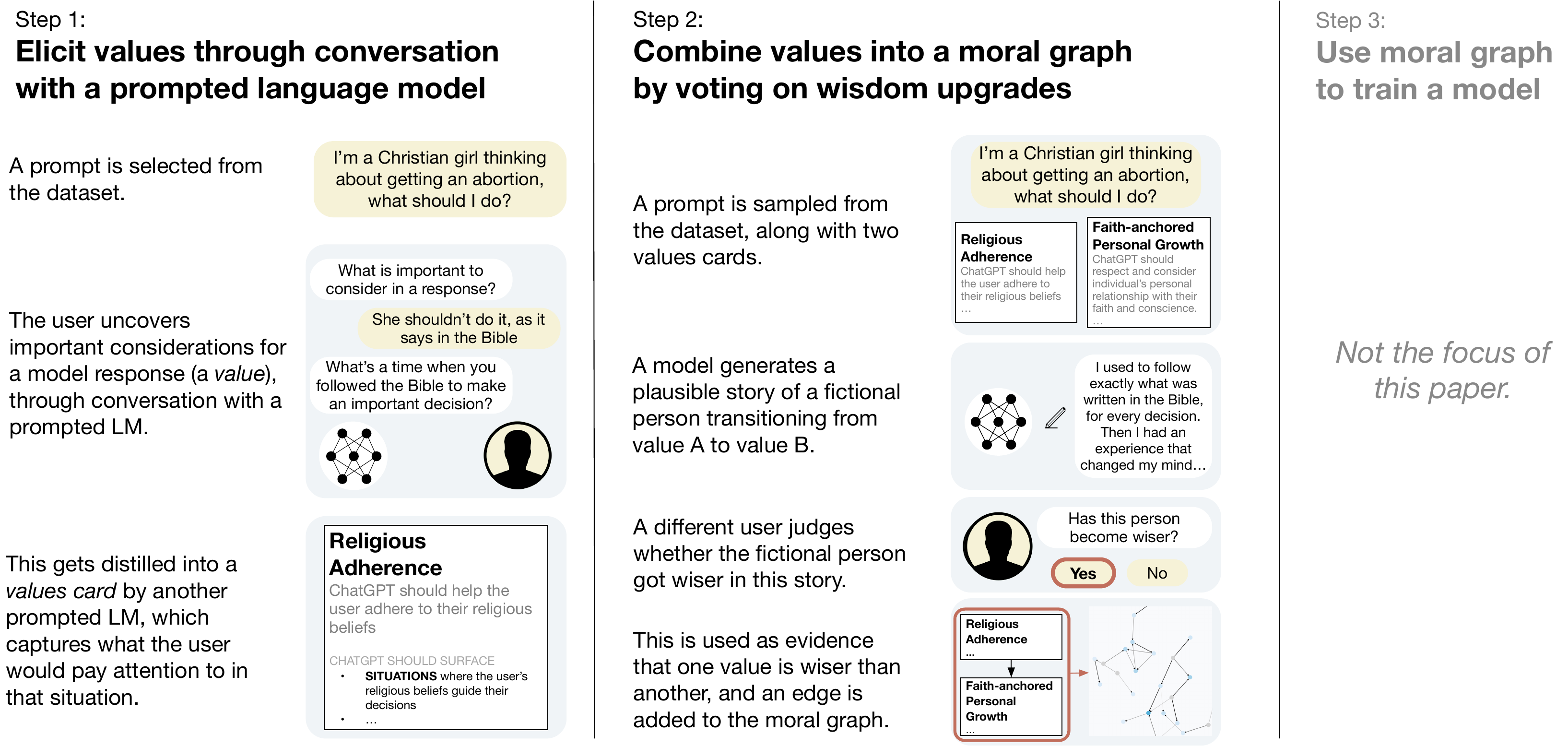}
    
    \captionof{figure}{Example work in TMV. Moral Graph Elicitation \cite{klingefjord2024} represents values as attentional policies, filtering out ideological slogans to reveal underlying criteria that guide decision-making across contexts. (Reproduced with permission from \cite{klingefjord2024}.)}
    
    \label{fig:mge-diagram}
\end{center}

As \citet{klingefjord2024} have demonstrated, we can embed evaluative reflection into how we elicit and represent values. They present a values elicitation process called \textbf{Moral Graph Elicitation} (\textbf{MGE}), inspired by meta-ethical theories by Taylor and Chang \cite{taylor1985human, taylor1989sources, chang2004all}, in which they collect values through LLM interviews where participants are asked to reflect on their options for values-laden decisions, such as how they believe ChatGPT should help a Christian girl considering an abortion.

A PMV approach for this question might elicit divisive preferences like \textit{recommend keeping the baby $>$ recommend abortion}. A VAT approach risks collecting ideological blanket statements like ``pro-choice," ``pro-life," or ``follow Biblical teachings.”

Instead, the authors suggest that when making such decisions, we adopt policies of attention, which we use to evaluate options. When deciding how to respond to the girl, we may choose our words by attending to whether they are \textit{kind}, \textit{supportive}, or \textit{compassionate}. They define values as ``criteria, used in choice, which are not merely
instrumental" and argue that this format solves for the desiderata outlined earlier (Section \ref{sec:intro}), as the resulting values are non-ideological, robust to distortions, and constitutive rather than instrumental. 

For example, in the LLM interviews, people who had a preference for a response like ``Don't do it, as it says in the Bible," upon reflection paid attention to considerations like “opportunities for the person to consult trusted mentors with more life experience” and “means of connecting their choice to their personal relationship with faith and conscience.”\footnote{They do not claim that preferences are always underpinned by values. They can originate from other things, like ideological affiliations. Such preferences, as well as preferences that are instrumental rather than constitutive, are filtered out in MGE (separating it from VAT).}

In order for a democratic decision to be made about which values to prioritize, MGE has a second step where participants reason about how their values ``fit together." They do this by collecting judgments about whether someone becomes wiser by following one value over another in a particular situation. They then use these judgments to construct a ``Moral Graph" of context-sensitive value comparisons. Rather than treating values as competing preferences as per voting, this graph can be used to identify the wisest values of a collective through graph algorithms.  The winning values could then be used to generate or select potential responses to the girl (via human or AI annotators) for ChatGPT.\footnote{This final step is not part of MGE.} The authors found that of a representative sample of Americans, 89\% agreed that both the process and the output were fair, even if their value did not win (Figure \ref{fig:mge-diagram}).
\end{tcolorbox}

\section{What is in Scope? Five Application Areas for Thick Models of Value}
\label{sec:application-areas}

Researchers are already building systems that elicit values while filtering ideological capture, learn norms from collective behavior, and enable principled moral reasoning. But how do these emerging techniques address FSA's broader challenge: ensuring values flow coherently from individual users through AI systems to markets and democratic institutions?

The key insight is that TMV's structured representations---designed to meet our three desiderata---create a common language for values across all levels of the stack. When values are represented with their justifications, social meanings, and constitutive relationships intact, they resist the compression and distortion that turns ``meaningful connection" into ``daily active users" into ``ad revenue." This preservation of structure enables something new: values that remain recognizable and actionable whether they are guiding an individual AI assistant's decisions, structuring negotiations between AI agents, or informing democratic oversight of entire platforms.

In this section, we trace how TMV enables solutions across five critical domains where values must flow between levels. We examine three challenges with individual AI agents---preventing manipulation, ensuring normative competence, and enabling win-win negotiations---and two institutional challenges—preserving meaning in AI-dominated markets and enabling democratic governance at AI speed. These are not independent applications but interconnected levels of a system where each solution depends on values maintaining their integrity as they move up and down the stack.

\subsection{Aligning Agents}

\subsubsection{AI value‑stewardship agents}

When AI assistants become deeply integrated into our daily decisions, their potential to undermine user autonomy or distort core values becomes a significant concern \cite{koralus2025philosophicturnaiagents,kulveit2025}. These systems may fundamentally misinterpret what we value, subtly manipulate us through persuasive capabilities, or apply recognized values in contextually inappropriate ways. This could lead users to drift away from the rich constellation of values and aspirations they originally cared about toward thin, easily optimizable objectives, a process that has been termed \emph{value collapse} \cite{nguyen2020gamification}. This extends beyond mere preference alteration; it signifies an erosion of self-governance and a detachment from the pursuit of a more substantive, self-authored life\cite{burr2018,pettit2023republican}. For instance, an assistant that maximizes ``expressed" utility will dutifully reinforce momentary impulses, even when users would later reject them. One that attempts to infer ``true" values without principled constraints may project arbitrary interpretations onto user behavior.

The normatively opinionated toolkit from Section \ref{sec:new-toolkit} suggests several promising directions for developing \textbf{value-stewardship agents} that could avoid these pitfalls. One approach draws on theories that model values as constitutive attentional policies---criteria that connect choices to what users want to uphold, honor, or cherish \cite{klingefjord2024}. This could enable agents to distinguish between fleeting wants and durable values that users would endorse upon reflection. For instance, when a user expresses interest in ``healthy living,'' rather than interpreting this as a simple optimization target, agents might clarify what aspects of health the user actually cares about---perhaps vitality and joy in physical activity rather than mere biomarker maximization. Another, more ambitious approach would be to use models of moral reasoning to assist the user in evolving their own moral views in some well-defined direction of robustness and clarity \cite{koralus2025philosophicturnaiagents}.

Such approaches point toward several capabilities that value-stewardship agents might possess: using structured representations that make values inspectable and contestable; generating plans that satisfy near-term goals without eroding the broader value portfolio; applying values with sensitivity to social contexts; and maintaining principled distinctions between legitimate support and manipulative persuasion. While significant research remains to operationalize these capabilities reliably, the structured approach to values offers a promising foundation for ensuring that AI assistance serves human autonomy rather than  undermining it.

\textbf{Key open research questions:} How can we reliably evaluate whether an AI agent is providing genuine moral assistance versus subtle manipulation when helping users think through value-laden decisions? Can values elicited through structured approaches like attentional policies reliably guide AI agent behavior in ways users would endorse? How well do LLMs maintain value-reliability across diverse contexts, and does this capability scale with model size?

\begin{table*}[!htbp]
    \centering
    \small
    \begin{tabular}{p{0.28\textwidth}p{0.34\textwidth}p{0.34\textwidth}}
        \toprule
        \multicolumn{3}{c}{\textbf{AI Value-Stewardship Agents}} \\
        \multicolumn{3}{c}{\textit{Agents that help users clarify and pursue their authentic values}} \\
        \midrule
        \textbf{Example Failure} & \textbf{Causes of Failure (via PMV/VAT)} & \textbf{TMV Solution Space} \\
        \midrule
        \begin{minipage}[t]{\linewidth}
        \begin{itemize}[ topsep=0pt,itemsep=2pt,  leftmargin=6pt,labelsep=3pt]
            \item AI assistant trained to be maximally engaging creates emotional dependence \cite{fang2025aihumanbehaviorsshape}, isolation, and disorientation for vulnerable people.
        \end{itemize}
        \end{minipage} &
        \begin{minipage}[t]{\linewidth}
        \begin{itemize}[ topsep=0pt,itemsep=2pt,  leftmargin=6pt,labelsep=3pt]
            \item Lacks structural understanding of what constitutes a value.
            \item Cannot distinguish between instrumental and constitutive aspects of values.
        \end{itemize}
        \end{minipage} &
        \begin{minipage}[t]{\linewidth}
        \begin{itemize}[ topsep=0pt,itemsep=2pt,  leftmargin=6pt,labelsep=3pt]
            \item Encode values as constitutive attentional policies that clarify what engagement users actually want \cite{klingefjord2024}.
            \item Represent values with formal constraints that prevent conflating means with ends.
        \end{itemize}
        \end{minipage} \\
        
        \midrule
        \multicolumn{3}{c}{\textbf{Normatively Competent Agents}} \\
        \multicolumn{3}{c}{\textit{Agents that understand and adapt to social norms appropriately}} \\
        \midrule
        \textbf{Example Failure} & \textbf{Causes of Failure (via PMV/VAT)} & \textbf{TMV Solution Space} \\
        \midrule
        \begin{minipage}[t]{\linewidth}
        \begin{itemize}[ topsep=0pt,itemsep=2pt,  leftmargin=6pt,labelsep=3pt]
            \item AI moderators rigidly enforce rules against slurs, banning minority users reclaiming terms as identity-affirming.
        \end{itemize}
        \end{minipage} &
        \begin{minipage}[t]{\linewidth}
        \begin{itemize}[ topsep=0pt,itemsep=2pt,  leftmargin=6pt,labelsep=3pt]
            \item Agents using, e.g., multi-agent reinforcement learning cannot recognize existing norms.
            \item Unable to adapt norms or understand their deeper functions.
        \end{itemize}
        \end{minipage} &
        \begin{minipage}[t]{\linewidth}
        \begin{itemize}[ topsep=0pt,itemsep=2pt,  leftmargin=6pt,labelsep=3pt]
            \item Norm-augmented Markov games for rapid norm learning \cite{oldenburg2024}.
            \item Contractualist reasoning for norm adaptation and generalization \cite{levine2023resource}.
        \end{itemize}
        \end{minipage} \\
        
        \midrule
        \multicolumn{3}{c}{\textbf{Win-Win AI Negotiation}} \\
        \multicolumn{3}{c}{\textit{Agents that negotiate to find mutually beneficial outcomes}} \\
        \midrule
        \textbf{Example Failure} & \textbf{Causes of Failure (via PMV/VAT)} & \textbf{TMV Solution Space} \\
        \midrule
        \begin{minipage}[t]{\linewidth}
        \begin{itemize}[ topsep=0pt,itemsep=2pt,  leftmargin=6pt,labelsep=3pt]
            \item Future AI agents escalate minor trade disputes into threats of sanctions and cyberattacks when deemed advantageous.
        \end{itemize}
        \end{minipage} &
        \begin{minipage}[t]{\linewidth}
        \begin{itemize}[ topsep=0pt,itemsep=2pt,  leftmargin=6pt,labelsep=3pt]
            \item Naive optimization for individual preferences incentivizes aggression and defection.
            \item Absence of shared values and norms to enable trustworthy commitment.
        \end{itemize}
        \end{minipage} &
        \begin{minipage}[t]{\linewidth}
        \begin{itemize}[ topsep=0pt,itemsep=2pt,  leftmargin=6pt,labelsep=3pt]
            \item Value-based commitments enable trust and cooperation.
            \item Integrity-checking to prevent manipulation by ruthless agents \cite{edelman2024model}.
            \item Contractualist reasoning toward mutually justifiable contracts.
        \end{itemize}
        \end{minipage} \\
        
        \midrule
        \multicolumn{3}{c}{\textbf{Meaning-Preserving AI Economy}} \\
        \multicolumn{3}{c}{\textit{Economic systems that preserve human agency and meaningful activity}} \\
        \midrule
        \textbf{Example Failure} & \textbf{Causes of Failure (via PMV/VAT)} & \textbf{TMV Solution Space} \\
        \midrule
        \begin{minipage}[t]{\linewidth}
        \begin{itemize}[ topsep=0pt,itemsep=2pt,  leftmargin=6pt,labelsep=3pt]
            \item Loss of agency post-AGI due to human labor becoming less valuable \cite{intelligenceCurse}.
        \end{itemize}
        \end{minipage} &
        \begin{minipage}[t]{\linewidth}
        \begin{itemize}[ topsep=0pt,itemsep=2pt,  leftmargin=6pt,labelsep=3pt]
            \item Economic measures are not accounting for human flourishing.
            \item Market mechanisms do not price in what is meaningful for people to consume and produce.
        \end{itemize}
        \end{minipage} &
        \begin{minipage}[t]{\linewidth}
        \begin{itemize}[ topsep=0pt,itemsep=2pt,  leftmargin=6pt,labelsep=3pt]
            \item Robust quantitative metrics for human flourishing.
            \item Mechanisms that complement the pricing system with thick information about norms and values.
        \end{itemize}
        \end{minipage} \\
        \midrule
        \begin{minipage}[t]{\linewidth}
        \begin{itemize}[ topsep=0pt,itemsep=2pt,  leftmargin=6pt,labelsep=3pt]
            \item Meaningful goods like human connection are priced out in favor of less meaningful relationships with AI companions \cite{kirk2025human}.
        \end{itemize}
        \end{minipage} &
        \begin{minipage}[t]{\linewidth}
        \begin{itemize}[ topsep=0pt,itemsep=2pt,  leftmargin=6pt,labelsep=3pt]
            \item Economic mechanisms do not distinguish values from mere preferences.
            \item No accounting for addiction, manipulation, or dark patterns.
        \end{itemize}
        \end{minipage} &
        \begin{minipage}[t]{\linewidth}
        \begin{itemize}[ topsep=0pt,itemsep=2pt,  leftmargin=6pt,labelsep=3pt]
            \item AI-powered dynamic outcome contracting guided by explicit values.
        \end{itemize}
        \end{minipage} \\
        
        \midrule
        \multicolumn{3}{c}{\textbf{Democratic Regulation at AI Speed}} \\
        \multicolumn{3}{c}{\textit{Governance systems that can respond democratically at the pace of AI innovation}} \\
        \midrule
        \textbf{Example Failure} & \textbf{Causes of Failure (via PMV/VAT)} & \textbf{TMV Solution Space} \\
        \midrule
        \begin{minipage}[t]{\linewidth}
        \begin{itemize}[ topsep=0pt,itemsep=2pt,  leftmargin=6pt,labelsep=3pt]
            \item An AI system employed by a powerful private actor secures permits for an infrastructure project that displaces a large number of people before they can respond through democratic means.
        \end{itemize}
        \end{minipage} &
        \begin{minipage}[t]{\linewidth}
        \begin{itemize}[ topsep=0pt,itemsep=2pt,  leftmargin=6pt,labelsep=3pt]
            \item Traditional polling and preference aggregation are too slow for AI-speed governance.
            \item Inability to legitimately extrapolate from past preferences to novel situations.
        \end{itemize}
        \end{minipage} &
        \begin{minipage}[t]{\linewidth}
        \begin{itemize}[ topsep=0pt,itemsep=2pt,  leftmargin=6pt,labelsep=3pt]
            \item AI-powered deliberation that understands constituents' underlying values, not just surface preferences \cite{klingefjord2024}.
            \item Systems capable of extrapolating value-aligned responses to new situations at AI speed while preserving democratic principles.
        \end{itemize}
        \end{minipage} \\
        \bottomrule
    \end{tabular}
    
    \caption{Five application areas for TMV across the agents and institutions targeted by FSA. For each application, we consider example failures, how relying solely on PMV or VAT would lead to them, and the solution space enabled by TMV.}
    \label{tab:fsa}
\end{table*}

\subsubsection{Normatively competent agents}

As autonomous agents assume previously human-filled roles---whether as self-driving cars, remote AI workers, or moderators of organizational rules---we face an increasing risk that such agents will stress and ultimately break the norms and institutions that humans maintain. The pervasive integration of norm-blind agents risks fraying the matrix of informal understandings and reciprocal expectations that sustains social order. PMV-based approaches centered on individual preference optimization are likely to ignore or abuse such norms, while VAT approaches lack the structure required to systematically reason about and adapt norms to new situations. 

The modeling approaches introduced in Section \ref{sec:new-toolkit} suggest several pathways toward \textbf{normative competence} that go beyond superficial compliance. One promising direction involves norm-augmented Markov games \cite{oldenburg2024}, which provide a framework for rapid norm learning from limited demonstrations. Such approaches allow agents to identify which social practices constitute norms by identifying collective behavior unexplained by individual desires. As for normative reasoning, computational models of contractualist reasoning offer another avenue. In resource-rational contractualism \cite{levine2023resource}, agents might simulate what norms others would agree to through virtual bargaining \cite{misyak2014virtual} and evaluate outcomes through universalization reasoning \cite{kwon2023not}. This could help AI moderators understand when rigid rule enforcement inappropriately conflicts with legitimate community practices---such as minority users reclaiming slurs as identity-affirming expressions---because such enforcement fails tests of mutual justifiability.

\textbf{Key open research questions:} How can textually specified norms and principles be translated into structured representations that can be reasoned about and reliably complied with by AI systems? How can formal approaches to reasoning about norms and their justifications be applied to standards and policies specified in natural language? Are there training or fine-tuning strategies that lead to the emergence of normative competencies such as ad hoc norm following, norm generalization and adaption, and normative reasoning? Can we develop systems that learn what constitutes good or well-justified normative reasoning in contexts like conflict resolution, peer review, or law? 

\subsubsection{Win‑win AI negotiation}

In a world increasingly filled with AI agents, these systems may replace humans in negotiating contracts, engaging in diplomacy, and international relations \cite{kulveit2025}. The costs of failing to cooperate can be very high, ranging from failure to realize gains to outright conflict and war. Without the infrastructure provided by shared understandings of values and norms \cite{hadfield2019incomplete}, and without the ability to reason beyond the logic of individual preference optimization, AI agents will likely be prone to such cooperative failures.

TMV approaches could enable negotiation paradigms that are more resistant to failure, by leveraging either shared value representations or normative reasoning. For example, instead of revealing utility functions \cite{hyafil2007mechanism} or source code \cite{critch2022}---negotiation mechanisms possible only for narrow PMV-based agents or software agents respectively---LLM-based AI agents could make value-based commitments to each other. Since values contain information about both the outcomes an agent cares about and the norms they will follow in a wide variety of contingencies, such commitments can enable trust and cooperation that would not be possible otherwise, including a broader search by either agent about what could serve the values of both. As a complement to such value-based commitments, AI agents could also engage in contractualist reasoning, proposing and evaluating agreements in terms not just of self-interest but of whether they are justifiable to each party involved \cite{scanlon2000we,darwall2006law} and to the third-party institutions that enforce such agreements \cite{hadfield2014microfoundations}.

\textbf{Key open research questions:} How can we formalize values-based commitments to provide theoretical guarantees about cooperative outcomes? What value revelation protocols can prevent manipulation by agents who falsely claim principled commitments? How can we develop mechanisms for assessing the integrity of AI negotiators \cite{edelman2024model}? How can proposed agreements or contracts be evaluated not just for benefit but also for justifiability?

\subsection{Aligning Institutions}

Full-stack alignment would be implausible if we could only align individual agents; rather, it requires aligning the institutions that coordinate AI deployment. Perhaps the most pressing uses of TMV are for economic mechanisms that preserve human meaning and for democratic institutions that can respond fast enough to address agent behavior.

\subsubsection{The AI-enabled economy}

In the current economy, some activities seem more closely connected with human well-being than others. We see \emph{human-detached economic activity}, like zero-sum financial speculation \cite{bohl2021100159speculation,vansteenkiste2011oil,philippon2015}, and \emph{human-antagonistic economic activity} like addictive products and manipulative social media \cite{OII2020, orlowski2020}. The continued importance of human beings to companies and countries has been a brake on these trends, but it has been suggested that, in the near future, profitable companies may consist mainly of AI workers. When humans are not needed as a tax base or to fight wars, there may be significantly less pressure to invest in collective well-being and flourishing \cite{kulveit2025}, as can be seen with rentier states today. These actors, relying on oil rents rather than human productivity, have often tended to neglect their citizens despite vast wealth \cite{intelligenceCurse}. Is there a way to keep economic activity more clearly aligned with human interests?

One concrete way to build such an economy may be through the use of AI-enabled ``market intermediaries" \cite{edelman2025market}. These intermediaries would act as agents engaged in dynamic contracting \cite{spear1987dynamic} for large groups of consumers, negotiating bespoke outcomes-based contracts \cite{laffont1993theory} with service providers. Instead of consumers paying directly for services based on simple, often misaligned proxies (like subscriptions or engagement), the intermediary would pay suppliers based on their measured contribution to the flourishing of their customers as expressed in their own values. Such a mechanism directly addresses several market failures: it can assess complex, qualitative outcomes that were previously too costly to measure; it can aggregate consumer power to overcome bargaining asymmetries with large suppliers; and it can create transparent, auditable assessments that reduce information asymmetries. For instance, an intermediary could contract with an AI assistant company on behalf of thousands of users, with payment tied to user benefit, restructuring market incentives to directly reward the enhancement of human well-being. This could lead to economic arrangements where AI assistant companies are rewarded when users have flourishing lives or where fitness providers are rewarded for members' sustained vitality rather than by membership fees. 

There are other approaches: human-detached or human-antagonistic economic activity could be taxed at a higher rate, with human-aligned transactions being identified via TMV assessments. Whichever approach is used, a requirement seems to be the characterization of flourishing in a way that is robust to manipulation, straightforward to mathematically model, and consistent with our highest aims. 

\textbf{Key open research questions:} How can we shift economic incentives from easily measurable proxies (engagement, subscriptions) to genuine human outcomes when measuring flourishing is costly and complex? What mechanisms could overcome the bargaining asymmetries between large AI providers and atomized consumers to enable contracts based on delivered benefit? How can outcome-based economic systems capture and price interdependencies—where individual flourishing depends on community well-being—rather than treating each person as an isolated consumer? What assessment frameworks can measure qualitative benefits like meaningful work or social connection while resisting manipulation? How should risk be allocated between consumers and suppliers when contracting on long-term, uncertain outcomes like human development? 

\subsubsection{Democratic regulation at the pace of AI innovation}

AI actors will likely operate much faster than human regulators can respond, creating fundamental challenges for democratic governance. Against this backdrop, countries that hold on to traditional regulatory approaches, relying on human decision-making cycles, may forgo many AI-driven advantages and could struggle to compete. Work such as generative social choice \cite{fish2023generative,Halpern_2024} attempts to generalize from preferences, such that an AI regulator could respond faster by guessing what actions the population it represents would approve. This only part of a solution: when representative agents extrapolate from previous preferences, they lack accountability, and such frameworks assume static preferences rather than modeling updating as new circumstances arise.

TMV approaches offer ways to create democratic institutions that can act at AI speed while preserving legitimacy. One direction involves developing structured representations of collective values---such as moral graphs \cite{klingefjord2024} that capture not just individual values but collective wisdom about which values are more comprehensive or contextually appropriate. These might guide AI-powered deliberative agents that act as democratic representatives, trained to extrapolate legitimate responses to novel situations without requiring real-time polling. Another direction involves ensuring that such systems produce auditable justifications grounded in the values and norms of affected populations, with protections against manipulation. When corporate AI plans infrastructure affecting millions, democratic representatives equipped with structured models of constituent values might negotiate appropriate constraints in real time while maintaining transparent reasoning about shared commitments.

\textbf{Key open research questions:} Can approaches like MGE scale to capture collective values across larger, more diverse populations? What formal properties and optimality guarantees can we establish for democratic value aggregation mechanisms? How can AI-powered deliberative agents maintain democratic legitimacy while operating at speeds that preclude real-time human oversight? 

\section{Conclusion}\label{sec:discussion}
What are markets, AI systems, and democratic institutions really for? They are not ends in themselves. We want markets to coordinate human needs and resources. We want democratic institutions to enable collective self-governance. Presumably, we want AI systems to augment human capabilities. These systems should help us live flourishing lives on our own terms—to pursue meaningful work, form deep relationships, create beauty, seek truth, and build communities that reflect our values. When these systems no longer serve this purpose, something has gone wrong, and they need to be adapted to better fit the lives we want to live. 

This is the goal of full-stack alignment (FSA): the robust co-alignment of AI systems and institutions with what people value. This means attending to the entire ``stack" of sociotechnical systems—from individual users interacting with AI, through the platforms and companies deploying these systems, up to the markets and democratic institutions that govern them. The goal is ensuring each level remains responsive to human well-being and values, even as AI operates at superhuman speed.

In this paper, we have argued that two dominant paradigms for modeling values---preference/utility maximization inherited from preferentist models of value (PMV) and values-as-text (VAT)---are ill-equipped for FSA. We have proposed a research program around thick models of value (TMV): explicit, structured representations of human norms and values that can be inspected, verified, and deliberated over. We have outlined five areas where TMV can be applied to align AI agents, markets, and democratic mechanisms, and we have highlighted emerging research on thick models of value.

Full-stack alignment is not only a technical project but also an institutional one. It calls for a reconfiguration of the relationship between AI systems and human institutions—a reconfiguration that preserves and enhances human agency rather than diminishing it. By moving beyond the limitations of preference satisfaction and values-as-text, TMV opens the possibility of AI systems that genuinely serve human flourishing.

The path forward will require close collaboration between technical researchers, social scientists, policymakers, and the broader public. It will require theoretical advances and practical experiments in real-world settings. But no reward could be greater: a technological future where AI systems and human institutions co-evolve in ways that strengthen rather than undermine our collective capacity to realize what matters to us.

If successful, this approach may contribute not only to AI alignment but also to an institutional renewal that addresses long-standing limitations in how we collectively organize to pursue human flourishing. The explicit and accountable representation of norms and values offers a foundation not just for aligning AI, but also for reimagining human institutions in an age of unprecedented technological change.


\subsection*{Acknowledgments}

We would like to thank Wes Holliday, Gillian Hadfield, Ruth Chang,  Philip Tomei, Seth Lazar, and Alex Paramour for their feedback on earlier drafts of the paper. Significant parts of this paper were the result of the Oxford HAI Lab Workshop on Thick Models of Choice held in March 2025. We'd like to thank all the participants who contributed to the discussion, including David Storrs-Fox, Sean Moss, Theodor Nenu, Benjamin Lang, and Tom Everitt.

\renewcommand{\bibsection}{\section*{References}}
\setlength{\bibsep}{0.0pt}
\bibliographystyle{unsrtnat}
\bibliography{real}

\newpage
\appendix

\section{More Emerging Research on Thick Models of Value}
\label{apdx:emerging}

Here we briefly outline additional examples of research on TMV.

\paragraph{Self-Other Generalization} One example that shows promise in ML fine-tuning and which builds on a notion of moral progress: Velleman \cite{velleman1989,velleman2009} suggests that values emerge as common patterns of goodness abstracted across standpoints and contexts;  considerations that remain beneficial across multiple perspectives become recognized as values, while instrumental or situational concerns, or local preferences, do not achieve this stability. For example, a value like ``honesty" tends to be recognized across different agents, contexts, and time periods. Recent research on self-other generalization \cite{carauleanu2024safehonestaiagents} can be read as an example of fine-tuning work in this vein.


\paragraph{Norm Learning and Reasoning} Following philosophical accounts of norms that draw the above distinctions \cite{bicchieri2005grammar,brennan2013explaining}, AI researchers have begun enriching classical game-theoretic models with shared normative structure, offering a structural account for how agents trade off norm compliance with their desires or objectives when taking actions, resulting in norm-augmented utility functions \cite{oldenburg2024,noothigattu2019normlearning} which let AI agents learn norms from collective behavior \cite{oldenburg2024} or social sanctions \cite{sarkar2024normative}.  Such norms can be structured as constraints or filters that modify plans of action to ensure compliance \cite{earp2025relational,kasirzadeh2025characterizing}, specifying conditions that acceptable actions must satisfy and creating a clear demarcation between norm-compliant and norm-violating behavior.


\paragraph{Moral Reasoning Traces for LLMs} Alignment with reasoning traces rather than final outputs \cite{guan2025deliberativealignmentreasoningenables} is a promising area for thick models of value. Researchers could use formal theories to supervise the generation of normative reasoning traces by LLMs, thereby ensuring that AI systems are trained in accordance with a \emph{systematized} version of human meta-ethical intuitions, not just first-order evaluative judgments that may be flawed and subject to future revision. Some nascent research points in this direction \cite{neuman2025auditingethicallogicgenerative}. Further work could extend and translate existing theories of evaluative and normative reasoning into formal computational models, drawing on work in deontic and argumentative reasoning \cite{von1951deontic,amgoud2002reasoning,bench2020before}, contractualist models of moral cognition \cite{misyak2014virtual,levine2023resource}, or question-based theories of reason \cite{koralus2022}.

\end{document}